\documentclass[conference]{IEEEtran}

\usepackage[cmex10]{amsmath}
\usepackage{amssymb}

\usepackage{algpseudocode}
\usepackage[caption=false, font=footnotesize]{subfig}
\usepackage{float}

\usepackage{ifpdf}
\ifpdf
	\usepackage[pdftex]{graphicx}
	\graphicspath{{figure-pdf/}}
	\DeclareGraphicsExtensions{.pdf, .jpg, .png, .tif}
\else
	\usepackage[dvips]{graphicx}
	\graphicspath{{figure-eps/}}
	\DeclareGraphicsExtensions{.eps, .ps}
\fi
\usepackage[bookmarks=false]{hyperref}
\usepackage{algorithm, balance, arydshln}
\IEEEoverridecommandlockouts

\begin{document}

\pdfpagewidth  8.50in
\pdfpageheight 11.00in

% --------------------------------------------------------------------
% The MOST number(s) is/are valid from August 1, 2015 to July 31, 2016.
% Please remind ypchen to update this number after July 31, 2016.
% --------------------------------------------------------------------
\newif\ifMOST
\MOSTtrue
%\MOSTfalse
\newcommand{\NCLMOSTNoOne}{MOST 104-2221-E-009-126}
%\newcommand{\NCLMOSTNoTwo}{}
% --------------------------------------------------------------------

% --------------------------------------------------------------------
% So it begins
% --------------------------------------------------------------------

%\author{
%\IEEEauthorblockN{Pei-Chuan~Tsai}
%\IEEEauthorblockA{Department of Computer Science\\
%National Chiao Tung University\\
%HsinChu City, TAIWAN\\
%Email: pctsai@nclab.tw}
%\and
%\IEEEauthorblockN{Chih-Ming~Chen}
%\IEEEauthorblockA{Department of Computer Science\\
%National Chiao Tung University\\
%HsinChu City, TAIWAN\\
%Email: ccming@nclab.tw}
%\and
%\IEEEauthorblockN{Ying-ping~Chen}
%\IEEEauthorblockA{Department of Computer Science\\
%National Chiao Tung University\\
%HsinChu City, TAIWAN\\
%Email: ypchen@cs.nctu.edu.tw}}

\newcommand{\coqTactic}[1]{\texttt{#1}}

\title{Automatically Proving Mathematical Theorems with Evolutionary Algorithms and Proof Assistants}
%\title{Automatically Proving Mathematical Theorems}
\author{
\IEEEauthorblockN{Li-An~Yang, Jui-Pin~Liu, Chao-Hong~Chen, and Ying-ping~Chen$^{*}$}
\IEEEauthorblockA{Department of Computer Science, National Chiao Tung University, HsinChu City, TAIWAN\\
layang@nclab.tw, jpliu@nclab.tw, chchen@nclab.tw, ypchen@cs.nctu.edu.tw}
}

%% paper title: Must keep \ \\ \LARGE\bf in it to leave enough margin.
%\title{\ \\ \LARGE\bf PSO-Based Evacuation Simulation Framework%
%	\thanks{Pei-Chuan~Tsai, Chih-Ming~Chen, and Ying-ping~Chen (corresponding author) are with the %Department of Computer Science, National Chiao Tung University, Hsinchu, TAIWAN. (e-mail: \{pctsai, %ccming, ypchen\}@nclab.tw).}}

%\author{Pei-Chuan~Tsai, Chih-Ming~Chen, and Ying-ping~Chen, \textit{Member, IEEE}}

\maketitle

% --------------------------------------------------------------------

\begin{abstract}

Mathematical theorems are human knowledge able to be accumulated in the form of symbolic representation, and proving theorems has been considered intelligent behavior. Based on the BHK interpretation and the Curry-Howard isomorphism, proof assistants, software capable of interacting with human for constructing formal proofs, have been developed in the past several decades. Since proofs can be considered and expressed as programs, proof assistants simplify and verify a proof by computationally evaluating the program corresponding to the proof. Thanks to the transformation from logic to computation, it is now possible to generate or search for formal proofs directly in the realm of computation. Evolutionary algorithms, known to be flexible and versatile, have been successfully applied to handle a variety of scientific and engineering problems in numerous disciplines for also several decades. Examining the feasibility of establishing the link between evolutionary algorithms, as the program generator, and proof assistants, as the proof verifier, in order to automatically find formal proofs to a given logic sentence is the primary goal of this study. In the article, we describe in detail our first, ad-hoc attempt to fully automatically prove theorems as well as the preliminary results. Ten simple theorems from various branches of mathematics were proven, and most of these theorems cannot be proven by using the tactic \coqTactic{auto} alone in Coq, the adopted proof assistant. The implication and potential influence of this study are discussed, and the developed source code with the obtained experimental results are released as open source.\par

\end{abstract}

% --------------------------------------------------------------------


% --------------------------------------------------------------------

\section{Introduction}
\label{sec:Introuduction}

Human knowledge has been accumulated in various forms. Among them are theorems in mathematics that are presented in a precise fashion and can be applied to innumerous domains. The importance of mathematics and its influence on science, engineering, and all kinds of technologies are undoubtedly beyond discussion. Modern mathematics are formal systems composed of four components: an alphabet, a grammar, axioms, and inference rules. Given the alphabet and grammar, in an information technology way of thinking, the development of a mathematical field or branch can be viewed as the growth of a database containing knowledge, in the form of logical sentences considered true or proven to be true. Initially, the database is empty. The first step is to put in the axioms, which are logical sentences considered true, followed by a loop: proving a new logical sentence by applying the inference rules to the database and inserting the proven logical sentence back to the database. The proven logical sentences are called theorems, propositions, lemmas, or corollaries.\par

Before the proposal of the link between logic and computation, the principle of Propositions as Types, logic and computation were previously considered two separate fields~\cite{Wadler:Propositions_as_Types}. Based on the BHK interpretation~\cite{BHK:book, wiki:BHK-interpretation} and the Curry-Howard isomorphism~\cite{CH:PNAS, CH:book, wiki:CH-isomorphism}, (functional) programming languages~\cite{FP:Why}, including Haskell~\cite{Hudak:2007:HHL:1238844.1238856, Haskell:official_site}, and proof assistants, such as Coq~\cite{Coq:Manual, Coq:official_site}, HOL~\cite{Nipkow-Paulson-Wenzel:2002, HOL:official_site}, Isabelle~\cite{Nipkow-Paulson-Wenzel:2002, Isabelle:official_site}, and LEGO~\cite{PollackPhD, LEGO:official_site} have been developed. Due to the transformation from logic to computation, proofs to mathematical theorems can then be expressed as programs and verified computationally.\par

Mathematical theorem proving has been considered intelligent behavior~\cite{Intelligent_behavior}. Even for today, while a Go computer program, called \textit{AlphaGo}, able to beat the human Go champion of Europe, Fan Hui~\cite{wiki:FanHui}, by five games to zero~\cite{Go:nature, Go:nature:2, Go:DeepMind} and Lee Sedol~\cite{wiki:LeeSedol}, who was ranked world \#2, by four games to one exists, the ability of computers to fully automatically prove mathematical theorems still seems quite limited. It is probably because most of the effort made on proof assistants aims at automated proof checking, which ensures the correctness of the proof, and at providing a proof development environment interacting with human. Moreover, there seems to be a lack of effective search mechanisms, which can ``think outside of the box,'' especially randomized or stochastic ones, used for this purpose.\par

Evolutionary algorithms~\cite{Back:1996:EAT:229867}, as stochastic methods, have been known for their flexility and versatility. They have been successfully applied to handle a variety of scientific and engineering problems in numerous disciplines. Hence, in this study, we would like to make our first attempt to link evolutionary algorithms, as the program generator, and proof assistants, as the proof verifier, in a simplistic way. The primary goal of this study is to assess the feasibility of establishing frameworks capable of fully automatically proving mathematical theorems by integrating the facility of generating programs and the facility of verifying proofs as evaluating programs. Specifically, we will design a straightforward evolutionary algorithm without complicated mechanisms and simply adopt Coq~\cite{Coq:official_site} for fitness evaluation. The preliminary results indicate the feasibility and demonstrate that this direction of research is promising.\par

The remainder of the paper is organized as follows. Section~\ref{sec:Background} introduces the background regarding the primary goal of the present work. Section~\ref{sec:Ad-Hoc_Attempt} describes in detail the ad-hoc attempt we adopt to investigate the feasibility of automatically proving mathematical theorems. The preliminary results obtained in our experiments are presented in section~\ref{sec:Preliminary_Results}, followed by section~\ref{sec:Discussion} giving a broad discussion on this work and its potential influence. Finally, section~\ref{sec:Conclusion} concludes the paper.\par

\section{Backound}
\label{sec:Background}

The goal of this study is to make an attempt to automatically prove mathematical theorems, and the present work is done by linking two well developed fundations, proof assistants and evolutionary algorithms. In this section, the background regarding the primiary goal of this study is presented.\par

First of all, \textit{automated theorem proving} is a term quite closely related to this study. Although automated theorem proving and the research line of this study may share the same ultimate goal, the ways to progress towards the ultimate goal are actually entirely different, especially that evolutionary algorithms, as stochastic search methods, are introduced in the present work to conduct global search to find formal proofs.\par

The term automated theorem proving appeared in the 1950s as the most developed field within automated reasoning~\cite{wiki:Automated_reasoning} and was applied in 1956 to Logic Theory Machine~\cite{LTM:1, Newell:1957:EEL:1455567.1455605}, a deduction system for the propositional logic which adopts a heuristic approach to emulate human reasoning. It was the first program designed to prove mathematical theorems. Another field of automated reasoning with less automation but more pragmatic actions is called \textit{interactive theorem proving}, in which proof assistants~\cite{wiki:Proof_assistant} have been developed. As software tools, proof assistants incorporate automated reasoning techniques to make logical decisions on mathematical theorems.\par

Proof assistants, such as Coq~\cite{Coq:official_site}, HOL~\cite{HOL:official_site}, Isabelle~\cite{Isabelle:official_site}, and LEGO~\cite{LEGO:official_site} have been designed to interact with users for developing formal proofs and to verify the logical validity of proofs. Among them is Coq, an interactive theorem prover developed by Inria~\cite{Inria:official_site} that allows users to express mathematical assertions by specifying various strategies called \textit{tactics}. It works based on the theory of the Calculus of Inductive Construction, a derivative of the calculus of constructions. Coq also provides a formal language to write mathematical definitions, executable algorithms and theorems together with an environment for semi-interactive development of machine-checked proofs. To help users to deal with formal proofs, Coq checks the validity of tactics step by step automatically. The features, capabilities, and representation of formalization of Coq are the reasons for Coq to be adopted in this study.\par

The other foundation is evolutionary algorithms~\cite{Back:1996:EAT:229867}, which have been successfully applied to resolve issues of many different natures in a host of domains. Evolutionary algorithms are known for their flexibility and versatility because their ability to handle black-box optimization makes it possible, even easy, to interface with nearly all kinds of search and optimization problems. Therefore, evolutionary algorithms are clearly suitable techniques currently available to be incorporated in the present work.\par

\section{Ad-Hoc Attempt}
\label{sec:Ad-Hoc_Attempt}

First of all, the source code developed in this study, the supporting materials required for conducting the experiments, and our experimental results are open source in the repository of nclab~\cite{nclab:GitHub} on GitHub~\cite{GitHub:official_site}. It should be viable and relatively easy for practitioners and interested readers to replicate the experimental results described later in this article.\par

Our present ad-hoc attempt, as suggested by the title of this paper, consists of two major components: an evolutionary algorithm for finding proofs and the proof assistant for verifying proofs. For part of the proof assistant, we employ Coq~\cite{Coq:official_site} in this study. The reason to employ Coq is quite straightforward that according to the entry of proof assistants~\cite{wiki:Proof_assistant} in Wikipedia, Coq supports most features, including higher-order logic and dependent types, which are most likely necessary if research along this line is continually pursued in the future.\par

\begin{figure}[t!]
\centerline{\includegraphics[width=80mm]{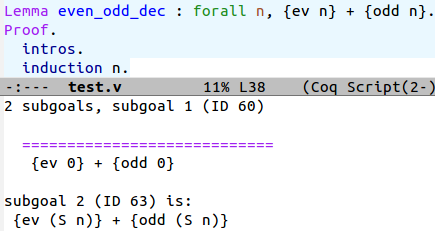}}
\caption{The figure shows that Coq helps a user to prove theorems in a step-by-step manner with tactics, which can be regarded as (CPU) instructions.}
\label{fig:EVEN_ODD_DEC}
\vspace{-2mm}
\end{figure}

Proof development in Coq is done through a sequence of tactics to interact with the current goals shown in the Coq interface. These tactics convert a goal of a proof into subgoals by implementing backward reasoning from conclusions to premises. For example, in order to prove $A \wedge B$, $A$ and $B$ must be both proven. Figure~\ref{fig:EVEN_ODD_DEC} shows a situation in which a Coq user strives to solve and eliminate subgoals stated below the line to prove the lemma or, programmatically speaking, to define the type of the lemma. Such a user-guided proof process allows certain degree of freedom and helps to find proofs in a procedural manner.\par

Because a deep structure of three-layered correspondence between logic and computation~\cite{Wadler:Propositions_as_Types} exists: propositions as types, proofs as programs, and simplification of proofs as evaluation of programs, Coq as a proof assistant has been developed based on this correspondence in order to provide the capability of checking the validity of proofs computationally. From the side of Coq's current intended usage, the logic side, Coq users give their proofs in Coq's way such that the proofs can be verified. Now, it is time to use Coq from the viewpoint of the other side, the programming side.\par

In some sense, we can consider that what Coq does is to provide an execution environment in which programs are written in the programming language defined by Coq. The tactics for composing proofs in Coq can be considered as instructions, such as those of x86 CPUs~\cite{wiki:x86_CPU} or ARM CPUs~\cite{wiki:ARM_CPU}. Thus, a proof written as a sequence of tactics in Coq can be viewed as a program composed of a sequence of (CPU) instructions. If the ``Coq CPU'' can successfully execute a given sequence of instructions, the proof expressed as this sequence of tactics is then verified to be logically valid. It is exactly what ``simplification of proofs as evaluation of programs~\cite{Wadler:Propositions_as_Types}'' indicates.\par

Thanks to the subtle and strong correspondence between proofs and programs, finding proofs can now be regarded as writing programs. In the realm of evolutionary computation, Genetic Programming~\cite{GP:1, GP:2, GP:3, GP:4} is a well known and established branch which aims at automatically generating computer programs fulfilling given specifications. Since Coq tactics are similar to CPU instructions, among the variants of Genetic Programming is Linear Genetic Programming~\cite{LGP:1, LGP:2} which is in particular designed for handling programs in the form of instruction sequences.\par

\renewcommand{\algorithmicensure}{\textbf{Output:}}

\begin{algorithm}[t!]
\caption{Flow of the adopted evolutionary algorithm}
\begin{algorithmic}[1]
\Procedure{Ad-Hoc\_EA}{$PopSize$, $MaxGen$}
	\State $t \leftarrow 0$;
    \State $Pop(t) \leftarrow \text{Initialization}(PopSize)$;
    \State $\text{Evaluation}(Pop(t))$;
	\Repeat
        \State $t' \leftarrow t+1$;
        \State $Pop(t') \leftarrow \emptyset$;
        \State $i \leftarrow 0$;
        \Repeat
            \State $p_1 \leftarrow \text{SelectOneParent}(Pop(t))$;
            \State $p_2 \leftarrow \text{SelectOneParent}(Pop(t))$;
            \State $c \leftarrow \text{Crossover}(p_1, p_2)$;
    		\If {$\text{UniformReal}[0, 1] \le MutRat$}
                \State $c \leftarrow \text{Mutation}(c)$;
    		\EndIf
            \State $Pop(t') \leftarrow Pop(t') \cup \{c\}$;
            \State $i \leftarrow i+1$;
        \Until {$i \ge PopSize$};
        \State $Pop(t) \leftarrow Pop(t')$;
            \Comment  Generational model
        \State $\text{Evaluation}(Pop(t))$;
        \State $t \leftarrow t'$;
	\Until {$t \ge MaxGen$};
\EndProcedure
\end{algorithmic}
\label{alg:Overview}
\end{algorithm}

While techniques developed for Linear Genetic Programming may be a good choice to perform the role of generating CPU instruction sequences, i.e., searching for proofs, we employ a more rudimentary evolutionary approach in this study, because the goal is to make a first, simplistic attempt for linking two domains, proof assistants and evolutionary algorithms, to examine whether some proof-of-principle results can be obtained. Search methods which are highly customized for generating programs of specific ``machines/CPUs'' might not be totally adequate for the ``Coq CPU,'' which is yet fully understood from the viewpoint of evolutionary algorithms.\par

\begin{algorithm}[t!]
\caption{Initialization of the population}
\begin{algorithmic}[1]
\Ensure
    Initialized population;
\Procedure{Initialization}{$PopSize$}
    \State $Pop \leftarrow \emptyset$;
    \State $i \leftarrow 0$;
    \Repeat
        \State $c \leftarrow ()$;
        \State $L \leftarrow \text{UniformInt}[L_{\ell}, L_{u}]$;
        \State $j \leftarrow 0$;
        \Repeat
            \State $T_{no} \leftarrow \text{UniformInt}[0, T_{max}]$;
            \State $\text{Append}(c, T_{no})$;
            \State $j \leftarrow j+1$;
        \Until {$j \ge L$};
        \State $Pop \leftarrow Pop \cup \{c\}$;
        \State $i \leftarrow i+1$;
    \Until {$i \ge PopSize$};
    \State return $Pop$;
\EndProcedure
\end{algorithmic}
\label{alg:Initialization}
\end{algorithm}

As a consequence, in this study, we adopt a simple, GP-like evolutionary algorithm of which the pseudo code is outlined in Algorithm~\ref{alg:Overview} to work with Coq using the standard library and a ``tactic base'' consisting of 153 tactics, which is available for download from our GitHub repository. Since the goal of this work is to make an attempt to link evolutionary algorithms and proof assistants, instead of aiming to find a proof of some specific logical sentence, we use only the maximum generation as the stopping criterion. For the experiments conducted in this study, we use the same tactic base and set $PopSize = 1000$, $MaxGen = 100$, and $MutRat = 0.25$. As explained, the tactic base can be considered as the ``Coq CPU'' instruction set. As an example, if there are 8 tactics in the tactic base, we number them from 0 to 7 as
\begin{equation}
    \text{Tacticbase: } \{T_0, T_1, T_2, T_3, T_4, T_5, T_6, T_7\}    \; .    \nonumber
    \label{eqn:tactic_base}
\end{equation}
Then, the chromosome representation adopted in this work is a variable-length integer sequence because the number of tactics needed to complete a proof may vary in length. For example, if there are chromosomes $A$ and $B$:
\begin{equation}
    \begin{aligned}
        \text{Chromosome } A: & (5, 0, 1, 6) \\
        \text{Chromosome } B: & (1, 5, 5, 7, 3, 6, 4)    \; ,    \nonumber
    \end{aligned}
    \label{eqn:representation}
\end{equation}
these two chromosomes respectively correspond to the following Coq proofs:
\begin{center}
\begin{tabular}{l:l}
    \multicolumn{1}{c:}{Individual $A$} & \multicolumn{1}{c}{Individual $B$}   \\
    Proof.                  & Proof.\\
    \ \ \coqTactic{intros}. &  \ \ \coqTactic{intros}.\\
    \ \ $T_5$.              & \ \ $T_1$.\\
    \ \ $T_0$.              & \ \ $T_5$.\\
    \ \ $T_1$.              & \ \ $T_5$.\\
    \ \ $T_6$.              & \ \ $T_7$.\\
    Qed.                    & \ \ $T_3$.\\
& \ \ $T_6$.\\
& \ \ $T_4$.\\
& Qed.
\end{tabular}
\end{center}
The tactic \coqTactic{intros} is not included in the adopted tactic base and is inserted in the front of each individual.\par

The procedure for initializing the population is given in Algorithm~\ref{alg:Initialization}. Currently, the tactic base contains 153 tactics, and thus, $T_{max}$ is 152. We use $L_{\ell} = 4$ and $L_{u} = 15$ in all the conducted experiments.\par

Parental selection, crossover, and mutation are given in Algorithms~\ref{alg:ParentalSelection}, \ref{alg:Crossover} and \ref{alg:Mutation}, respectively. For a given population, we select two individuals among the top 50\% as parents to go through crossover and, probably, mutation to generate one offspring. The individuals ranked top 50\% have an equal probability to be selected as parents. The employed crossover operator is essentially the well known one-point crossover. Because the lengths of the two chromosomes are probably different, the cut point is uniformly chosen within the length of the shorter chromosome. The crossover creates only one child, and the child consists of the first parent before the cut point and the second parent after the cut point. The mutation operator may be applied with the mutation rate $MutRat$. One of the tactics of the individual to be mutated is replaced by one from the tactic base. Both choices are uniformly random.\par

\begin{algorithm}[t!]
\caption{Parental selection}
\begin{algorithmic}[1]
\Ensure
    One selected parent;
\Procedure{SelectOneParent}{$Pop$}
    \State $i \leftarrow \text{UniformInt}[0, PopSize/2)$;
        \State \Comment The 0th is the highest.
    \State return the individual with the $i$th highest fitness;
\EndProcedure
\end{algorithmic}
\label{alg:ParentalSelection}
\end{algorithm}

\begin{algorithm}[t!]
\caption{Crossover}
\begin{algorithmic}[1]
\Ensure
    One offspring generated by crossover;
\Procedure{Crossover}{$p_1, p_2$}
    \State $i \leftarrow \text{UniformInt}[1, \text{Min}(p_1.length, p_2.length)]$;
    \State $c \leftarrow p_1[0, i-1] + p_2[i, p2.length-1]$;
    \State return $c$;
\EndProcedure
\end{algorithmic}
\label{alg:Crossover}
\end{algorithm}

\begin{algorithm}[t!]
\caption{Mutation}
\begin{algorithmic}[1]
\Ensure
    One mutated invidual;
\Procedure{Mutation}{$c$}
    \State $i \leftarrow \text{UniformInt}[0, c.length-1]$;
    \State $c[i] \leftarrow \text{UniformInt}[0, T_{max}]$;
    \State return $c$;
\EndProcedure
\end{algorithmic}
\label{alg:Mutation}
\end{algorithm}

Finally, the evaluation procedure invokes Coq to evaluate individuals in the population and is shown in Algorithm~\ref{alg:Evaluation}. As aforementioned, this study is a proof-of-principle one to check the viability of linking evolutionary algorithms and proof assistants. We would like to consider Coq as a black box and make use of the results obtained by simply invoking Coq. Hence, we use ``the number of tactics successfully verified by Coq'' as the fitness value. Such fitness might actually be inappropriate because apparently, longer chromosomes are preferred to shorter ones, and it is somewhat counterintuitive to common mathematical sense. However, from the viewpoint of evolutionary algorithms, longer chromosomes in the population might play the role of a pool of partial proofs, and adopting such fitness may just be beneficial in this regard.\par

\vspace{1mm}

In order to use the number of tactics as fitness, some limitation is required to avoid meaningless conditions. For example, the tactic \coqTactic{simpl} may be repeated forever, and tactics regarding the commutative laws can be repeated forever by pairs. For this kind of cases, we give a list of unrepeatable tactics without affecting Coq's operations. In addition to the number of passed tactics, if Coq gives an error message saying ``No such unproven subgoal,'' we consider the proof is complete with extra redundant tactics and assign a fitness value of 1000 minus ``the number of tactics successfully verified by Coq,'' where 1000 is an arbitrary large number used to separate complete proofs and the others. For complete proofs, shorter proofs are still preferred because of higher scores.\par

\begin{algorithm}[t!]
\caption{Evaluation}
\begin{algorithmic}[1]
\Procedure{Evaluation}{$PopSize$}
    \State $i \leftarrow 0$;
    \Repeat
        \State Invoke Coq on individual $i$;
        \State $s \leftarrow$ the number of passed tactics;
		\If {Error: No such unproven subgoal}
            \State Assign ($1000-s$) as fitness to individual $i$;
        \Else
            \State Assign $s$ as fitness to individual $i$;
		\EndIf
		\State $i \leftarrow i+1$;
    \Until {$i \ge PopSize$};
\EndProcedure
\end{algorithmic}
\label{alg:Evaluation}
\end{algorithm}

In the source code available from our GitHub repository, we also pick some individuals with the top scores for further examination by invoking Coq on them in a tactic-by-tactic manner. If a complete proof is found, it is archived for our research purpose and nothing regarding the evolutionary process is changed. This step, or in fact using $MaxGen$ as the stopping criterion, may not be necessary if in the future, finding a proof to some given logical sentence is the only goal.\par

\section{Preliminary Results}
\label{sec:Preliminary_Results}

With the adopted evolutionary algorithm and our ad-hoc attempt presented in section~\ref{sec:Ad-Hoc_Attempt}, we have successfully proven ten theorems of different branches in mathematics automatically. As aforementioned, the proofs found by the introduced approach for the following theorems are also available as supporting materials in our GitHub repository.
\begin{itemize}
    \item {\bf Arithmetic}
        \begin{itemize}
            \item {\it Theorem n\_le\_k} : forall n k, n = 0 $\rightarrow$ n $\leq$ k.
            \item {\it Theorem plus\_n\_0} : forall n, n + 0 = 0.
            \item {\it Theorem n\_1\_n} : forall n, n \textasciicircum \ 1 = n.
        \end{itemize}
    \item {\bf Logic}
        \begin{itemize}
            \item {\it Lemma solving\_by\_eapply} : forall(P Q : nat $\rightarrow$ Prop), \\
                $\text{\ \ }$(forall n k, Q k $\rightarrow$ P n) $\rightarrow$ Q 1 $\rightarrow$ P 2.
            \item {\it Theorem andb\_prop} : forall n k, \\
                $\text{\ \ }$andb n k = true $\rightarrow$ n = true $\wedge$ k = true.
            \item {\it Theorem andb\_true\_elim2} : forall n k : bool, \\
                $\text{\ \ }$andb n k = true $\rightarrow$ k = true.
            \item {\it Theorem andb\_true\_intro} : forall n k, \\
                $\text{\ \ }$n = true $\wedge$ k = true $\rightarrow$ andb n k = true.
        \end{itemize}
    \item {\bf Parity}
        \begin{itemize}
            \item {\it Theorem ev\_minus2} : forall n, \\
                $\text{\ \ }$ev n $\rightarrow$ ev (pred (pred n)).
            \item {\it Theorem SSev\_even} : forall n, ev (S (S n)) $\rightarrow$ ev n.
            \item {\it Theorem silly\_prob} : \\
                $\text{\ \ }$(forall n, evenb n = true $\rightarrow$ oddb (S n) = true) \\
                $\text{\ \ }$$\rightarrow$ oddb 4 = true.
        \end{itemize}
\end{itemize}
Note that the tactic \coqTactic{auto} is not included in the tactic base, and more importantly, most of the theorems listed above cannot be proven by using only \coqTactic{auto} in Coq. Interested readers may further investigate into Coq and check out the supporting materials for more information.\par

The obtained, preliminary results are quite within our expectation. We were able to prove a number of simple theorems as listed above, while we failed to do the same on relatively advanced theorems. Simple theorems are more likely to be proven in several steps with different sequences of tactics. For a concrete example, we can consider one here:
\begin{equation}
    \begin{aligned}
    \text{{\it Theorem }} &\text{{\it andb\_true\_intro} : forall n k,} \\
    &\text{n = true $\wedge$ k = true $\rightarrow$ andb n k = true.}
    \end{aligned}
    \label{eqn:andb_true_intro}
\end{equation}
According to our experimental results, more than thirty different valid sequences of Coq tactics have been found in one hundred generations with 153 tactics in the tactic base. As aforementioned, the size of population, i.e., the number of individuals, is 1000, and in one successful run, the first complete proof was found at the 8th generation. Since the 8th generation, the evolutionary process continued to generate more than a few distinct sequences of tactics, although these seemingly different proofs might not be entirely mathematically different. Such a situation indicates that the diversity of proofs exists and will be further discussed in the next section.\par

Despite the fact that proven theorems in these preliminary results seem quite simple, even straightforward in human eyes, the potential for research and development on linking the two domains, evolutionary algorithms and proof assistants, can clearly be identified. These ten proven theorems are from various branches of mathematics, and the proofs are found by using the identical evolutionary algorithm with the same tactic base of Coq. Thus, the current attempt can be considered neutral, not biased towards certain branch of mathematics.\par

It can also be justified that finding the proofs to the above theorems is hardly by chance. The capability of automatically proving mathematical theorems does exist in a system linking together evolutionary algorithms and proof assistants. Taking the theorem of Equation~(\ref{eqn:andb_true_intro}) and 153 Coq tactics for example, one has barely no chance to come up with a valid proof using a pure random search. Because the first proof, whose length was 5, was discovered at the 8th generation with a population of 1000 individuals, the probability for a pure random search to find a proof with the same computational resource can be estimated approximately
\begin{equation}
\label{prob}
{\frac{8 \times 1000}{153^5}} \approx 9.542 \times 10^{-8} \; .
\end{equation}
If there are actually 100 valid proofs of different sequences of which the length is 5 or shorter than 5, the probability is still about the magnitude of $10^{-6}$ or $10^{-5}$. On the other hand, it took around only 8000 evaluations to ``guess it right,'' which costs little with modern computational facilities.\par

In summary, ten simple mathematical theorems from various branches of mathematics, most of them cannot be proven by the current, limited automatic mechanism of proof assistants alone, were proven by the proposed ad-hoc approach with a reasonable cost of computational resource. The feasibility of automatically generating complete proofs to mathematical theorems is clearly indicated.\par

\section{Discussion}
\label{sec:Discussion}

In order to assess the feasibility of automatically proving mathematical theorems, an ad-hoc attempt, composed of a straightforward evolutionary algorithm and the direct use of Coq, was described in section~\ref{sec:Ad-Hoc_Attempt}. The preliminary results presented in section~\ref{sec:Preliminary_Results} provide the evidence that the line of research on linking evolutionary algorithms and proof assistants is worth pursuing. Thus, the primary goal of this study has been achieved.\par

As mentioned in section~\ref{sec:Preliminary_Results}, the proposed ad-hoc approach failed to prove relatively advanced theorems which require proofs of longer tactic sequences. Although it is reasonable and acceptable in the present work, efforts on both sides need to be done to develop an automatic theorem proving framework for practical use.\par

Firstly, on the side of proof assistants, in the ad-hoc attempt, we basically use the number of passed tactics and its calculation as fitness. In addition to forbidding some tactics to repeat, we may consider different fitness assignment for compound usage of certain tactics or variable fitness weights on tactics for some sequence patterns. Moreover, instead of simply invoking Coq and directly using the returned results, looking into the running status of Coq may give essential information on the proving progress. If the internal (running) status of Coq can be obtained and utilized for evaluating individuals, the evolutionary process will be much better guided and the searching/proving ability of the system will be greatly enhanced.\par

On the side of the search methodology, since the feasibility to use evolutionary algorithms to search for formal proofs has been preliminarily identified, techniques from Linear Genetic Programming can then be considered and put into action. In addition to Linear Genetic Programming, other variants of Genetic Programming, such as Cartesian Genetic Programming~\cite{CGP:1, CGP:2, CGP:3}, or even techniques commonly used in Genetic Programming, such as automatically defined function (ADF) and automatically defined macro (ADM), can be integrated into the evolutionary proof-search engine to extract ``modularized'' segments of proofs in a sense which may yet be understandable for the time being.\par

Further into the realm of randomized search methods, given the recent success in Go playing with computer programs~\cite{Go:nature:2}, Monte Carlo tree search~\cite{MCTS:survey} may also be considered as the search mechanism. Although a great amount of efforts will be required for its adoption and adaptation in the theorem proving framework, Monte Carlo tree search is a promising methd alternative to the techniques of Genetic Programming.\par

From the obtained the preliminary results, we can easily find that even an automatic theorem proving framework as simple as our ad-hoc attempt is capable of finding alternative proofs, composed of different sequences of tactics, to a given theorem. Figures~\ref{fig:Thm1a} to \ref{fig:Thm2d} show some of these alternative proofs found in our experiments.\par

\newcommand{\ypTabularVSpacing}{\noalign{\vskip 2mm}}

\begin{figure}[t!]
\begin{center}
\begin{tabular}{c}
    {\it Theorem n\_le\_k} : forall n k, n = 0 $\rightarrow$ n $\leq$ k. \\
\ypTabularVSpacing
\begin{tabular}{l:l}
    Proof.                              & Proof.\\
    \ \ \coqTactic{intros}.             & \ \ \coqTactic{intros}.\\
    \ \ \coqTactic{rewrite H}.          & \ \ \coqTactic{inversion H}.\\
    \ \ \coqTactic{eapply le\_0\_n}.    & \ \ \coqTactic{eapply le\_0\_n}.\\
    Qed.                                & Qed.
\end{tabular}
\end{tabular}
\end{center}
\caption{To prove the theorem, the left proof finds the first subterm matching the hypothesis in the goal. The right proof inverts the quality of the hypothesis statement based on the constructor.}
\label{fig:Thm1a}
\end{figure}

\begin{figure}[t!]
\vspace{2mm}
\begin{center}
\begin{tabular}{c}
    {\it Theorem ev\_minus2} : forall n, ev n $\rightarrow$ ev (pred (pred n)). \\
\ypTabularVSpacing
\begin{tabular}{l:l}
    Proof.                          & Proof.\\
    \ \ \coqTactic{intros}.         & \ \ \coqTactic{intros}.\\
    \ \ \coqTactic{induction H}.    & \ \ \coqTactic{simpl}.\\
    \ \ \coqTactic{eapply ev\_0}.   & \ \ \coqTactic{inversion H}.\\
    \ \ \coqTactic{exact H}.        & \ \ \coqTactic{eapply ev\_0}.\\
    Qed.                            & \ \ \coqTactic{assumption}.\\
                                    & Qed.
\end{tabular}
\end{tabular}
\end{center}
\caption{Induction vs. Inversion}
\label{fig:Thm3a}
\end{figure}

\begin{figure}[t!]
\vspace{2mm}
\begin{center}
\begin{tabular}{c}
    {\it Theorem SSev\_even} : forall n, ev (S (S n)) $\rightarrow$ ev n. \\
\ypTabularVSpacing
\begin{tabular}{l:l}
    Proof.                          & Proof.\\
    \ \ \coqTactic{intros}.         & \ \ \coqTactic{intros}.\\
    \ \ \coqTactic{inversion H}.    & \ \ \coqTactic{inversion H}.\\
    \ \ \coqTactic{exact H1}.       & \ \ \coqTactic{induction H0}.\\
    Qed.                            & \ \ \coqTactic{assumption}.\\
                                    & Qed.
\end{tabular}
\end{tabular}
\end{center}
\caption{Inversion vs. Inversion + Induction}
\label{fig:Thm3b}
\end{figure}

These tactic sequences verified by Coq might not be totally different from the viewpoint of concepts in mathematics, while they still differ from each other with the distinct adoption of logical strategies expressed by the tactic sequences. Taking the proofs in Figure~\ref{fig:Thm2d} as an example, one can easily discriminate these proofs by examining whether or not the technique of Mathematical Induction is used. As a metaphor, we can think that these proofs reach the same intersection from different roads. Such an ability to provide diverse proofs is valuable and important, because the diversity either for an entire proof or in proof segments may not only help to automatically construct proofs to advanced theorems but also possibly inspire mathematicians to develop complicated proofs.\par

\begin{figure}[t!]
\begin{center}
\begin{tabular}{c}
     {\it Theorem n\_1\_n} : forall n, n \textasciicircum \ 1 = n. \\
\ypTabularVSpacing
\begin{tabular}{l:l}
    Proof.                                          & Proof.\\
    \ \ \coqTactic{intros}.                         & \ \ \coqTactic{intros}.\\
    \ \ \coqTactic{rewrite$\,$<-$\,$mult\_1\_l}.    & \ \ \coqTactic{rewrite$\,$<-$\,$mult\_1\_r}.\\
    \ \ \coqTactic{eapply mult\_comm}.              & \ \ \coqTactic{reflexivity}.\\
    Qed.                                            & Qed.\\
\end{tabular}\\
\ypTabularVSpacing \hdashline \ypTabularVSpacing
\begin{tabular}{l}
    Proof.\\
    \ \ \coqTactic{intros}.\\
    \ \ \coqTactic{induction n}.\\
    \ \ \coqTactic{trivial}.\\
    \ \ \coqTactic{rewrite$\,$<-$\,$mult\_1\_r}.\\
    \ \ \coqTactic{trivial}.\\
    Qed.
\end{tabular}
\end{tabular}
\end{center}
\caption{No induction vs. Induction}
\label{fig:Thm1c}
\end{figure}

\begin{figure}[t!]
\vspace{2mm}
\begin{center}
\begin{tabular}{c}
\begin{tabular}{l}
     {\it Theorem andb\_true\_intro} : forall n k, \\
        $\text{\ \ }$n = true $\wedge$ k = true $\rightarrow$ andb n k = true.
\end{tabular}\\
\ypTabularVSpacing
\begin{tabular}{l:l}
    Proof.                          & Proof.\\
    \ \ \coqTactic{intros}.         & \ \ \coqTactic{intros}.\\
    \ \ \coqTactic{inversion H}.    & \ \ \coqTactic{induction H}.\\
    \ \ \coqTactic{rewrite H0}.     & \ \ \coqTactic{induction n}.\\
    \ \ \coqTactic{trivial}.        & \ \ \coqTactic{exact H0}.\\
    Qed.                            & \ \ \coqTactic{eapply H}.\\
                                    & Qed.
\end{tabular}\\
\ypTabularVSpacing \hdashline \ypTabularVSpacing
\begin{tabular}{l}
    Proof.\\
    \ \ \coqTactic{intros}.\\
    \ \ \coqTactic{inversion H}.\\
    \ \ \coqTactic{induction n}.\\
    \ \ \coqTactic{eapply H1}.\\
    \ \ \coqTactic{rewrite H1}.\\
    \ \ \coqTactic{exact H0}.\\
    Qed.
\end{tabular}
\end{tabular}
\end{center}
\caption{Induction vs. Inversion vs. Both}
\label{fig:Thm2d}
\end{figure}

It must be noted that the nature and properties of this work are distinctly different from those of conventional optimization tasks handled by using mathematical or evolutionary optimization techniques, although proofs are indeed found via application of optimization methods. Unlike optimization tasks are often repeatedly solved, there is in fact no need to prove a known theorem once again. Hence, the future development of this work should focus on the capability of handling the structural flexibility of proofs in order to prove non-trivial theorems, instead of on the efficiency, either finding a proof faster or finding a shorter proof. Efficiency should not be the primary goal, at least not at early stages of this type of work. If in the future, finding shorter proofs is required, advanced techniques, such as superoptimization~\cite{Massalin:1987:SLS:36206.36194, Schkufza:2013:SS:2451116.2451150}, may be utilized.\par

The influence of this line of research may be profound, if frameworks ready for practical use can be developed. Mathematical theorems are human knowledge able to be accumulated in a database with its symbolic representation. Apparently, all the theorems known to human as of today were proven by mathematicians, professional or amateur, or genius. Ordinary people without proper training and specific talents probably have little chance to make contributions in this regard. However, if automatic theorem proving frameworks exist, with the support of certain online cloud services, proving mathematical theorems can be actually \textit{crowdsourced}. Imaging that if there is a database containing all the mathematical theorems ever proven by human, and any one with a computer can donate computational power for proving some unknown conjectures, human knowledge can then be accumulated in an unprecedented speed.\par

If we further stretch this probably already farfetched possibility, considering that a database containing all the proven logical sentences, i.e., theorems, exists, all the proofs to these theorems can be collected along the way of establishing this database. These proofs are also extremely useful and valuable information. As aforementioned, for a given theorem, alternative proofs may shed light on deep mathematical connections and possibly invoke mathematicians' intuition. On the other hand, if there are repeating or similar tactic sequence patterns in the proofs across different, perhaps unrelated theorems, novel proof techniques may be discovered and identified. Data mining methods and techniques proposed in data science can be applied to the collected proofs in order to further accelerate the acquiring of knowledge.\par

\section{Conclusions}
\label{sec:Conclusion}

In this paper, we introduced the possibility and potential to establish the link between evolutionary algorithms and proof assistants, both of which have been respectively developed in the past several decades while have not been linked to each other. The primary goal of this study was to assess the viability to automatically prove mathematical theorems. In order to achieve the goal in a simplistic way, we adopted Coq, one of the available, well developed proof assistants for validating proofs and designed a straightforward evolutionary algorithm capable of generating integer sequences, interpreted as Coq tactic sequences, i.e., proofs. The preliminary results indicated that the direction of linking evolutionary algorithms and proof assistants is worth pursuing, and the potential influence of the possible outcomes will be remarkably significant in many aspects of advancing and accumulating human knowledge.\par

As discussed in section~\ref{sec:Discussion}, the immediate future work includes adopting advanced techniques in existence, such as Linear Genetic Programming and Cartesian Genetic Programming, and investigating into Coq to utilize the Coq internal status at run time for better fitness assignment and therefore better evolutionary guidance. Moreover, Monte Carlo tree search can also be considered to enhance the search ability.\par

A reasonable first milestone for this line of research is to fully automatically prove non-trivial, well known theorems, such as Fermat's little theorem. At the stage, accumulating human knowledge on mathematics in a database of symbolic representation is viable by (re-)finding proofs to known theorems. The next milestone, which seems quite farfetched for the time being, is to prove existing mathematical conjectures. If such a milestone is reached, advancing human knowledge on mathematics can then be done in an automatic, computational way. Finally, the ultimate milestone is to prove unknown, meaningful theorems without human intervention, while in addition to the capability of automatically proving theorems, it undoubtedly requires other intelligent behavior and abilities, including creativity, speculation, and desire to solve problems.\par

% --------------------------------------------------------------------

\section*{Acknowledgments}
\label{sec:Acknowledgments}

\ifMOST
The work was supported in part by the Ministry of Science and Technology of Taiwan under Grant {\NCLMOSTNoOne}.
\fi
The authors are grateful to the National Center for High-performance Computing for computer time and facilities.\par

% --------------------------------------------------------------------

\balance

%\Urlmuskip=0mu plus 1mu\relax
\bibliographystyle{IEEEtran}
\bibliography{CEC2016_GPcoq}

% --------------------------------------------------------------------
% Document ends here
% --------------------------------------------------------------------
\end{document}